\documentclass[10pt,twocolumn,letterpaper]{article}
\PassOptionsToPackage{table}{xcolor}
\usepackage[OT1]{fontenc}
\usepackage[pagenumbers]{cvpr}
\usepackage{graphicx}
\usepackage{flushend}
\usepackage{placeins}

\definecolor{cvprblue}{rgb}{0.21,0.49,0.74}
\usepackage[pagebackref,breaklinks,colorlinks,allcolors=cvprblue,pdfversion=1.7]{hyperref}

\title{KineState-SWAM: A Surgical World Action Model with Explicit Kinematics and Causal Interaction Dynamics}
\author{%
  Rulin Zhou\textsuperscript{1,2},
  Yujie Ma\textsuperscript{3},
  Rui Tang\textsuperscript{4},
  Qiujie Song\textsuperscript{1,2},
  Wanhao Liu\textsuperscript{5}\\[2pt]
  Guoheng Ma\textsuperscript{6},
  Liangjing Shao\textsuperscript{1,2},
  An Wang\textsuperscript{1,2},
  Hongliang Ren\textsuperscript{1,2}\\[5pt]
  {\normalsize\textsuperscript{1}The Chinese University of Hong Kong\quad
   \textsuperscript{2}Shenzhen Loop Area Institute}\\
  {\normalsize\textsuperscript{3}Shenzhen University\quad
   \textsuperscript{4}The Hong Kong University of Science and Technology (Guangzhou)}\\
  {\normalsize\textsuperscript{5}Guangdong University of Technology\quad
   \textsuperscript{6}Southern University of Science and Technology}%
}
\hypersetup{%
  pdftitle={KineState-SWAM: A Surgical World Action Model with Explicit Kinematics and Causal Interaction Dynamics},
  pdfauthor={Rulin Zhou, Yujie Ma, Rui Tang, Qiujie Song, Wanhao Liu, Guoheng Ma, Liangjing Shao, An Wang, Hongliang Ren}
}

\begin{document}
\maketitle
\begin{abstract}
Existing surgical world models often leave implicit how instrument motion drives localized tissue responses and how interaction history informs subsequent actions. We introduce \textbf{KineState-SWAM}, a surgical world--action model coupling action prediction and video generation through a persistent interaction state, separating kinematically determined instrument motion from learned scene evolution. A flow-matching policy predicts joint trajectories conditioned on interaction state, robot configuration, and task instruction. \textbf{Kinematic Geometry Conditioning} converts these trajectories into surface-motion tokens and spatially aligned instrument renderings using calibrated URDF kinematics. \textbf{Causal Interaction Dynamics} recurrently updates the state through geometry-guided attention and motion-aligned feature transport, carrying interaction history into subsequent action predictions without intermediate video synthesis. The \textbf{State--Geometry Adapter} combines state-dependent spatial modulation with localized geometric residuals to guide video generation. Interaction-aware supervision emphasizes tissue responses near instrument working regions. We introduce \textbf{dVRK-Chole5.2k}, comprising 5,255 videos spanning 17 surgical tasks. Experiments demonstrate improved action accuracy and video fidelity over evaluated baselines within matched action-conditioning settings. Ablations support preserving and transporting interaction states for long-horizon prediction. 
\end{abstract}

\section{Introduction}
\label{sec:introduction}

\begin{figure*}[t]
    \centering
    \includegraphics[width=\textwidth]{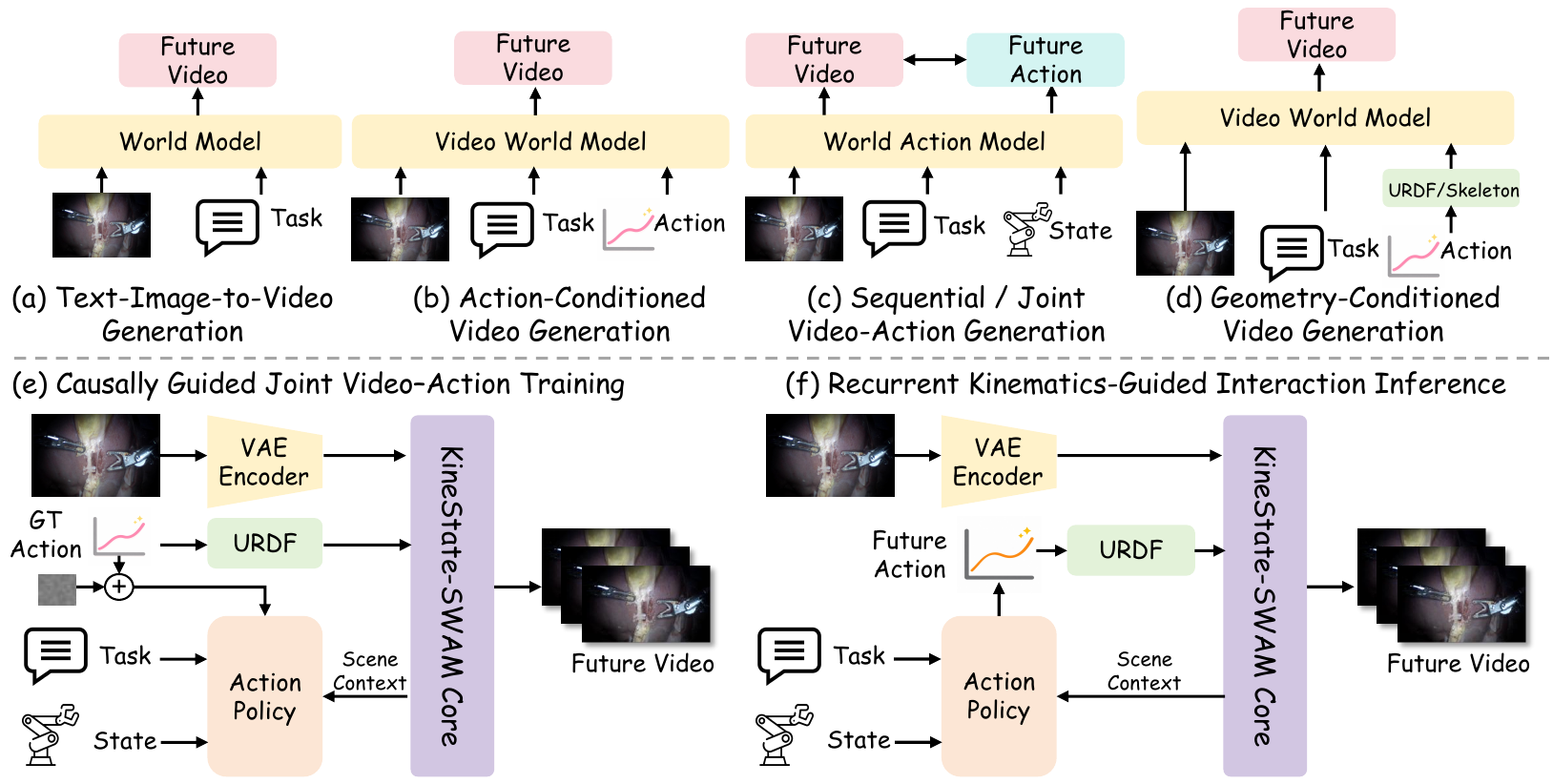}
\caption{Comparison of world-modeling paradigms and KineState-SWAM. Our model predicts robot actions and uses their URDF-derived motion to update a persistent interaction state, which guides subsequent action prediction and video generation.}
    \label{fig:overview}
      \vskip -0.1in
\end{figure*}

Learning to anticipate the consequences of robot actions is an important step toward surgical world models that support simulation and policy learning~\cite{he2025cosmos,tejero2026nvidia}. Such models must account for both articulated instrument motion and the evolution of deformable tissue. In manipulation tasks such as grasping, retracting, and dissecting, the visual consequences of an action depend on the ongoing instrument--tissue interaction. These consequences also shape the context in which subsequent actions are selected. A useful surgical world--action model should therefore connect action prediction to scene evolution while retaining interaction history, providing a consistent description of what the robot does and how the scene responds.

Recent video foundation models, including CogVideoX, HunyuanVideo-1.5, and Wan, provide rich visual and motion priors for text- and image-conditioned video synthesis (Figure~\ref{fig:overview}(a))~\cite{yang2025cogvideox,wu2025hunyuanvideo,wan2025wan}. Building on such priors, embodied world models connect visual prediction with robot actions. As illustrated in Figure~\ref{fig:overview}(b), Ctrl-World supports action-conditioned rollouts through frame-level conditioning and pose-conditioned memory retrieval~\cite{guo2026ctrl}. World--action models further integrate action generation (Figure~\ref{fig:overview}(c)): DreamZero jointly models future videos and actions, OpenWAM investigates modular world--action pretraining, and X-WAM extends joint prediction to RGB-D observations with asynchronous denoising~\cite{ye2026world,wang2026openwam,guo2026unified}. In the surgical domain, Endora explores generative endoscopy simulation, while Surgical Vision World Model learns latent actions from unlabeled surgical videos~\cite{li2024endora,koju2025surgical}. Cosmos-H-Surgical uses generated videos and inferred pseudokinematics to support policy learning, whereas Cosmos-H-Dreams enables real-time, kinematics-driven streaming simulation~\cite{he2025cosmos,tejero2026nvidia}. SurgVista further addresses tissue-deformation consistency and long-horizon prediction drift through deformation regularization and drift-adaptation training~\cite{pan2026surgvista}. Collectively, these advances extend video generation toward action-aware world modeling, motivating more explicit representations of robot motion and its influence on scene evolution.

A complementary line of research introduces spatially aligned visual and geometric conditions to control world-model predictions (Figure~\ref{fig:overview} (d)). DWM conditions interaction videos on scene and hand-mesh renderings, while GigaWorld-1 combines spatially aligned controls with hierarchical visual history for robot-policy evaluation~\cite{kim2026dexterous,team2026gigaworld}. Robot kinematics provides a direct route from joint trajectories to such visual conditions. BridgeV2W renders pixel-aligned embodiment masks from URDF models and camera parameters, Kinema4D projects kinematic robot trajectories into spatiotemporal pointmaps, and GeniWorld uses URDF-rendered visual actions to condition environmental dynamics~\cite{chen2026bridgev2w,xu2026kinema4d,gu2026geniworld}. Surgical methods develop related control interfaces tailored to instrument structure and scene appearance. Surg-UniWorld integrates edge, depth, and optical-flow conditions through modality experts anchored to first-frame appearance and hierarchical semantic masks~\cite{zhou2026surg}. SurgCast predicts future instrument skeletons from current structure, robot state, and prospective actions, then injects the predicted geometry and derived motion descriptors through geometric residuals and conditioning modulation~\cite{liu2026surgcast}. KVLR lifts articulated kinematics into image-aligned semantic, depth, orientation, velocity, and acceleration fields, routing these conditions across modalities and motion scales~\cite{li2026articulated}. These approaches establish explicit geometric interfaces for controllable surgical video generation. For surgical world--action modeling, we focus on a persistent interaction state that evolves under kinematic motion and provides shared context for both subsequent action prediction and video synthesis.

In surgical videos, instrument appearance and motion often occupy larger image regions than the localized tissue responses induced by robot actions. This spatial imbalance can bias global video reconstruction toward instrument fidelity, leaving interaction-induced tissue deformation underemphasized. Yet these local responses are important for understanding how the scene evolves under robot motion and for informing subsequent actions. Given joint trajectories and calibration, URDF kinematics can determine instrument geometry and motion analytically, whereas tissue responses must be inferred from the current scene and its interaction history. This distinction motivates using kinematically determined instrument motion as an explicit driver of learned scene evolution, allowing the model to focus on interaction-dependent tissue changes. A persistent interaction state, recurrently updated under this driver, retains the effects of preceding manipulations and provides shared, history-aware context for subsequent action prediction and scene evolution.

\begin{figure*}[t]
    \centering
    \includegraphics[width=\textwidth]{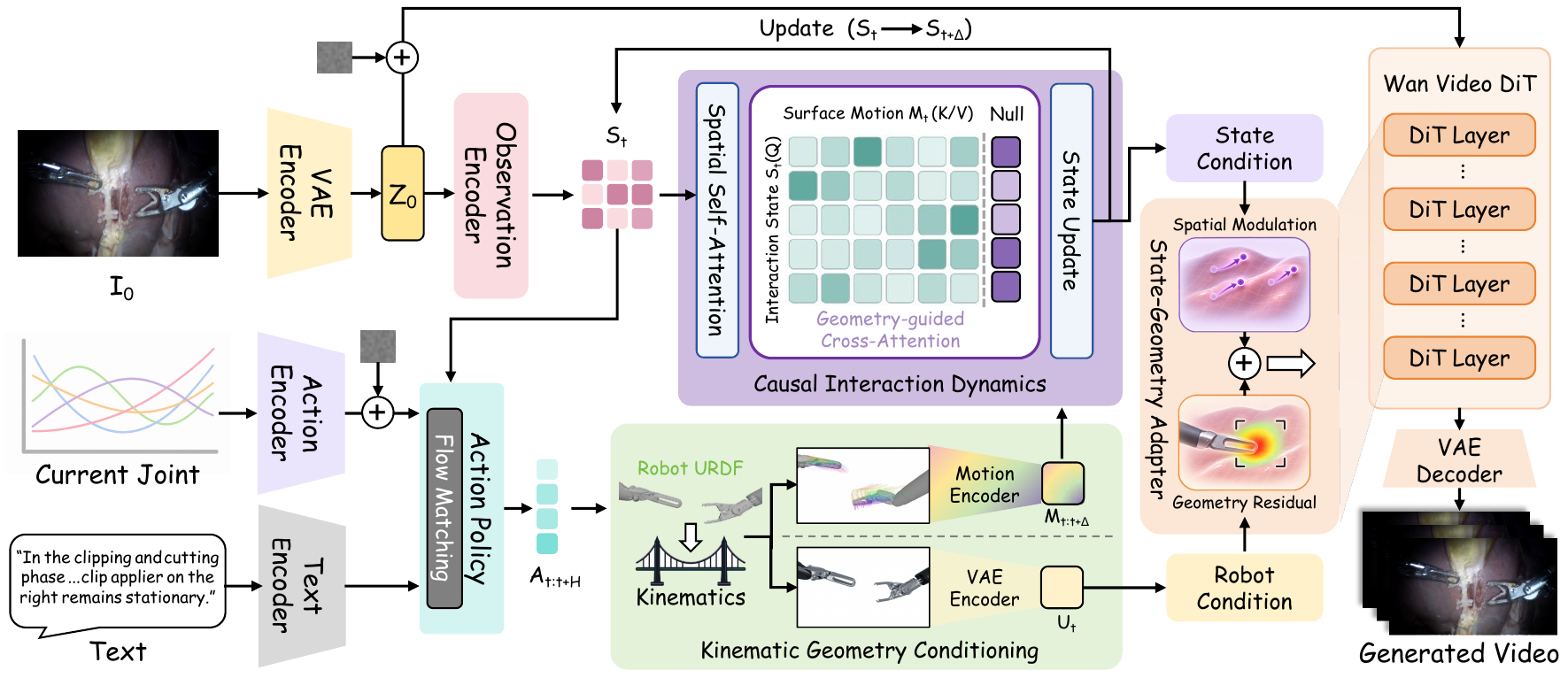}
   \caption{
   The flow-matching Action Policy predicts future robot joint trajectories, which are transformed into explicit instrument-motion and rendering conditions through
    calibrated URDF kinematics. Causal Interaction Dynamics recurrently updates a persistent interaction state, while the State--Geometry Adapter injects complementary state and geometric conditions into a pretrained video DiT.
}

    \label{fig:method}
         \vskip -0.1in
\end{figure*}

We propose \textbf{KineState-SWAM}, a surgical world--action model centered on a persistent interaction state shared by action prediction and video generation. The flow-matching Action Policy predicts an action chunk from this state, the current robot configuration, and the task instruction. Kinematic Geometry Conditioning uses calibrated URDF kinematics to embed joint trajectories into sparse motion tokens that retain surface correspondences and dense, spatially aligned features from instrument renderings. Causal Interaction Dynamics recurrently evolves the state through geometry-guided attention, motion-aligned feature transport, and gated residual updates, linking instrument motion to interaction-dependent scene changes. During training, ground-truth trajectories condition state evolution and video generation while the policy learns future action prediction. At inference, predicted instrument motion drives state updates within each action chunk, whose terminal state and final predicted joint configuration then condition the next policy call. This action--state recurrence preserves interaction history without requiring intermediate video synthesis. To generate video, the State--Geometry Adapter translates the evolving interaction context into spatial feature modulation, while localized geometric residuals guide instrument synthesis. To facilitate systematic evaluation of action and video prediction, we further introduce \textbf{dVRK-Chole5.2k}, comprising 5,255 videos spanning 17 surgical subtasks of robotic cholecystectomy. 
\section{Method}

\subsection{Problem Formulation and Overview}
\label{sec:formulation}

Given an initial surgical observation $I_0$, robot configuration $\mathbf{q}_0$, task instruction $\mathbf{c}$, and a calibrated robot URDF $\mathcal{U}$, we predict future joint trajectories and their corresponding surgical video. We maintain a persistent interaction state $\mathbf{S}_t$, initialized from the observation and instrument geometry, to connect kinematically determined motion with learned scene evolution and provide shared context for action and video prediction. As illustrated in Figure~\ref{fig:method}, each action chunk starting at $t$ follows:
\begin{equation}
\begin{aligned}
\hat{\mathbf{Q}}_{t+1:t+H}
&= \pi_{\theta}(\mathbf{S}_t,\mathbf{q}_t,\mathbf{c}),\\
\mathbf{M}_{\tau:\tau+\Delta}
&= \mathcal{K}_{\mathcal{U}}(\mathbf{q}_{\tau},\mathbf{Q}_{\tau+1:\tau+\Delta}^{\mathrm{cond}}),\\
\mathbf{S}_{\tau+\Delta}
&= \mathcal{F}_{\phi}(\mathbf{S}_{\tau},\mathbf{M}_{\tau:\tau+\Delta}),
\end{aligned}
\label{eq:framework}
\end{equation}
where $\pi_{\theta}$ denotes the Action Policy, $\mathcal{K}_{\mathcal{U}}$ extracts URDF-derived surface motion, and $\mathcal{F}_{\phi}$ represents Causal Interaction Dynamics. State updates proceed sequentially over $\tau\in\{t,t+\Delta,\ldots,t+H-\Delta\}$, following ground-truth trajectories during training and predicted trajectories during inference. With $H=16$ and $\Delta=4$, the entire predicted chunk drives four state updates before the policy replans from $\mathbf{S}_{t+16}$ and the final predicted configuration $\hat{\mathbf{q}}_{t+16}$. This action--state recurrence preserves interaction history without intermediate video synthesis. The evolving states and spatially aligned instrument renderings condition video generation through the State--Geometry Adapter.

\subsection{Action Prediction and Kinematic Geometry}
\label{sec:kinematics}

\noindent\textbf{Action prediction.} We employ a conditional flow-matching policy to predict future robot joint trajectories from the current interaction state $\mathbf{S}_t$, robot configuration $\mathbf{q}_t$, and task instruction $\mathbf{c}$. A Transformer-based velocity field models joint displacements relative to the current configuration, conditioned on interaction-state features, task instructions, and relative action timestamps. During inference, the learned velocity field progressively transforms Gaussian noise into joint displacements, which are converted into joint trajectories. The resulting trajectories provide explicit motion inputs for subsequent kinematic geometry conditioning.

\noindent\textbf{Kinematic geometry conditioning.} Accurate surgical scene prediction requires distinguishing deterministic instrument motion from interaction-dependent tissue responses. Rather than learning both implicitly from video, we exploit known robot kinematics to provide explicit geometric guidance for subsequent scene evolution. Given a joint trajectory $\mathbf{Q}^{\mathrm{cond}}$ and a calibrated robot URDF $\mathcal{U}$, we analytically recover articulated instrument motion through forward kinematics. Each canonical surface point $\bar{\mathbf{p}}_i$ is associated with a fixed instrument link $\ell_i$, allowing its camera-frame position at time $\tau$ as
\begin{equation}
\mathbf{p}_i^{\tau} = \Pi_3\left(\mathbf{T}_{C\leftarrow B}\mathbf{T}_{B\leftarrow\ell_i}(\mathbf{q}_{\tau};\mathcal{U})\tilde{\mathbf{p}}_i\right),
\label{eq:kinematic_projection}
\end{equation}
where $\tilde{\mathbf{p}}_i$ denotes the homogeneous canonical coordinate, $\mathbf{T}_{B\leftarrow\ell_i}$ is the link transformation, $\mathbf{T}_{C\leftarrow B}$ is the calibrated robot-to-camera transformation, and $\Pi_3$ converts homogeneous coordinates into 3D positions. The fixed link association preserves surface correspondences across articulated motion. As illustrated in Figure~\ref{fig:method}, this kinematic representation produces two complementary conditions: sparse surface-motion tokens for interaction-state evolution and dense rendered geometry for video generation.

\noindent\textbf{Surface motion encoding.} We sample canonical surface points from both instruments and encode their articulated motion over each state-update interval. For each point, we aggregate its link poses and motion descriptors together with canonical coordinates, relative timestamps, and arm/link embeddings:
\begin{equation}
\begin{aligned}
&\mathbf{h}_i^{\mathrm{kin}} = [\mathbf{h}_i^{\mathrm{pose}},\mathbf{h}_i^{\mathrm{motion}},\bar{\mathbf{p}}_i,\boldsymbol{\delta},\mathbf{e}_i],\\
&\mathbf{M}_{t:t'} = \left\{f_{\mathrm{mot}}(\mathbf{h}_i^{\mathrm{kin}})\right\}_{i=1}^{N_s},
\end{aligned}
\label{eq:surface_motion}
\end{equation}
where $f_{\mathrm{mot}}$ denotes a learnable motion encoder and $N_s$ is the number of sampled surface points. By preserving canonical surface identities, the resulting tokens capture articulated instrument motion without requiring visual correspondence estimation. These tokens serve as geometric keys and values in the subsequent Causal Interaction Dynamics module, explicitly conditioning scene evolution on instrument motion.

\noindent\textbf{Rendered geometry encoding.} We render the articulated URDF meshes along $\mathbf{Q}^{\mathrm{cond}}$ in the calibrated camera view and encode the resulting grayscale sequence using a frozen video VAE. The encoded features are spatially packed and concatenated with local geometric and motion descriptors and relative temporal offsets to form dense geometry features $\mathbf{U}$. We derive a spatial support map $\mathbf{P}_R$ by max-pooling the projected instrument occupancy to retain thin structures and applying spatial dilation to extend support to neighboring regions. The State--Geometry Adapter uses $\mathbf{U}$ and $\mathbf{P}_R$ to provide localized geometric guidance for video generation, complementing the surface-motion tokens used for interaction-state evolution.

\subsection{Causal Interaction Dynamics}
\label{sec:interaction}

\noindent\textbf{Interaction state initialization.} We initialize a persistent spatial interaction state $\mathbf{S}_0\in\mathbb{R}^{N\times d}$ from the initial surgical observation and instrument geometry. A frozen VAE extracts visual features, which are spatially packed into tokens $\mathbf{Z}_0=\operatorname{Pack}(\mathcal{E}_{\mathrm{VAE}}(I_0))$. These tokens are combined with projected instrument geometry and spatial positional embeddings:
\begin{equation}
\mathbf{S}_0=\left[\mathbf{Z}_0,\;f_g(\mathbf{G}_0)+\mathbf{E}_{xy}\right]+f_o(\mathbf{Z}_0),
\label{eq:state_initialization}
\end{equation}
where $\mathbf{G}_0$ contains initial instrument coverage and depth, $\mathbf{E}_{xy}$ denotes spatial positional embeddings, and $f_g$ and $f_o$ are learnable projections. This initialization preserves visual information from the observed scene while incorporating geometric context for subsequent state evolution.

\noindent\textbf{Geometry-guided interaction attention.} To model scene evolution under articulated instrument motion, we condition the interaction-state transition on URDF-derived surface-motion tokens $\mathbf{M}_{t:t'}$. As illustrated in Figure~\ref{fig:method}, spatial self-attention first aggregates scene context into $\mathbf{H}_t$, followed by geometry-guided cross-attention that uses scene features as queries and instrument-motion tokens as keys and values:
\begin{equation}
\begin{aligned}
&\mathbf{Q}_a=W_Q\mathbf{H}_t, \mathbf{K}_a=[W_K\mathbf{M}_{t:t'},\mathbf{k}_{\varnothing}],\\
&\mathbf{V}_a=[W_V\mathbf{M}_{t:t'},\mathbf{0}],\\
&\mathbf{A}_t=\operatorname{Softmax}\left(\frac{\mathbf{Q}_a\mathbf{K}_a^{\top}}{\sqrt{d}}+\mathbf{B}_{\mathrm{geo}}\right)\mathbf{V}_a,
\end{aligned}
\label{eq:kinematic_attention}
\end{equation}
where $\mathbf{B}_{\mathrm{geo}}$ encodes geometric distance biases between spatial state locations and instrument surfaces, and $\mathbf{k}_{\varnothing}$ denotes a learnable null key with zero-valued content. The null token allows each state query to selectively disregard irrelevant instrument motion. By directing kinematic information toward the scene representation, the interaction transformer incorporates motion-dependent cues while preserving spatial context from preceding interactions.

\noindent\textbf{Recurrent state update.} To preserve spatial information while capturing interaction-induced changes, we decompose state evolution into motion-aligned feature transport and learned residual correction. The interaction transformer predicts a latent displacement field $\mathbf{F}_t$, a residual update $\Delta\mathbf{S}_t$, and an update rate $\boldsymbol{\rho}_t$. The next interaction state is computed as
\begin{equation}
\begin{aligned}
&\widetilde{\mathbf{S}}_{t'}=\mathcal{W}(\mathbf{S}_t,\mathbf{F}_t),
\boldsymbol{\eta}_t=1-\exp\left(-\operatorname{softplus}(\boldsymbol{\rho}_t)\Delta\tau\right),\\
&\mathbf{S}_{t'}=\widetilde{\mathbf{S}}_{t'}+\boldsymbol{\eta}_t\odot\Delta\mathbf{S}_t,
\end{aligned}
\label{eq:interaction_transition}
\end{equation}
where $\mathcal{W}$ transports spatial features according to the learned displacement field, and $\Delta\tau$ is the elapsed time. Feature transport propagates existing scene information, while the gated residual captures interaction-dependent changes beyond spatial displacement. The displacement field operates entirely in latent space. The updated state is recursively carried across prediction windows, preserving interaction history for subsequent action prediction without intermediate video synthesis.

\subsection{State--Geometry-Conditioned\texorpdfstring{\hfil\break}{ }Video Generation}
\label{sec:video}

\noindent\textbf{Decoupled state--geometry conditioning.} The recurrent interaction state captures history-dependent scene evolution, while URDF-derived geometry provides explicit instrument configurations. Rather than merging these complementary representations into a single condition, we construct separate state and robot conditioning pathways, as illustrated in Figure~\ref{fig:method}. For each video latent timestamp $\tau$, the two conditions are defined as
\begin{equation}
\begin{aligned}
&\mathbf{C}_{S}^{\tau}=f_S([\mathbf{S}_{\tau},\mathbf{S}_{\tau}-\mathbf{S}_{\tau^-},\delta_\tau]),
&\mathbf{C}_{R}^{\tau}=f_R(\mathbf{U}_{\tau}),
\end{aligned}
\label{eq:video_conditions}
\end{equation}
where $\tau^-$ denotes the preceding state timestamp, $\delta_\tau$ is the temporal interval, and $f_S$ and $f_R$ are learnable projections. The state condition combines accumulated interaction context with temporal changes, while the robot condition encodes spatially aligned instrument geometry and motion. Both conditions are aligned with the video latent tokens and remain separate.

\noindent\textbf{State--Geometry Adapter.} We introduce a State--Geometry Adapter that integrates the two conditions through spatially varying state modulation and localized geometric residuals. Given the hidden features $\mathbf{H}_{\ell}$ of a video Transformer block, we obtain projected features $\mathbf{X}_{\ell}=W_p\operatorname{LN}(\mathbf{H}_{\ell})$. The state condition produces modulation parameters $[\boldsymbol{\gamma}_{\ell},\boldsymbol{\beta}_{\ell}]=f_m(\mathbf{C}_S+\mathbf{e}_{\ell})$, where $\mathbf{e}_{\ell}$ embeds the video noise level and layer identity. The two branches are formulated as
\begin{equation}
\begin{aligned}
&\mathbf{R}_{\ell}^{S}=f_{\mathrm{sp}}\left((1+\tanh\boldsymbol{\gamma}_{\ell})\odot\mathbf{X}_{\ell}+\boldsymbol{\beta}_{\ell}\right),\\
&\mathbf{R}_{\ell}^{R}=f_{\mathrm{pt}}\left([\mathbf{X}_{\ell},\mathbf{C}_R+\mathbf{e}_{\ell}]\right),
\end{aligned}
\label{eq:state_geometry_adapter}
\end{equation}
where $f_{\mathrm{sp}}$ employs depthwise spatiotemporal convolution and channel projection, while $f_{\mathrm{pt}}$ applies pointwise feature projection. The state branch modulates generative features across the scene, whereas the robot branch constructs geometry-conditioned residuals. To restrict direct geometric injection to instrument-related regions, we employ the URDF-derived support map $\mathbf{P}_R$:
\begin{equation}
\begin{aligned}
&\mathbf{D}_{\ell}=\mathbf{R}_{\ell}^{S}+\mathbf{P}_{R}\odot\mathbf{R}_{\ell}^{R},
&\mathbf{H}_{\ell}^{+}=\mathbf{H}_{\ell}+\mathbf{m}_{\mathrm{fut}}\odot W_{\ell}(\mathbf{D}_{\ell}),
\end{aligned}
\label{eq:adapter_injection}
\end{equation}
where $\mathbf{m}_{\mathrm{fut}}$ restricts residual injection to future video tokens, and $W_{\ell}$ is a zero-initialized output projection. This design allows interaction states to guide spatially varying scene evolution while explicitly conditioning instrument regions on their kinematic geometry, without indiscriminately applying geometric residuals across the scene.

\noindent\textbf{Conditional video generation.} We incorporate the proposed adapter into selected layers of a pretrained Wan Video DiT without modifying its attention architecture. The zero-initialized projections preserve the original backbone behavior at initialization. During flow-matching sampling, the conditioning boundary latent remains fixed, while future video latents are progressively generated under the state and geometry conditions. The resulting latents are decoded into RGB frames using the frozen VAE decoder.

\subsection{Training Objectives}
\label{sec:training}

We optimize action and video prediction with separate flow-matching objectives. During training, ground-truth joint trajectories drive interaction-state evolution and condition video generation, while recurrent states are supervised through RGB reconstruction.

To emphasize localized tissue responses, we construct a kinematics-guided interaction prior $A$ by projecting swept instrument working regions into the video latent space. Given the per-token flow-matching error $e$ and environment mask $E$, we define
\begin{equation}
\begin{aligned}
\mathcal{L}_{\mathrm{env}}^{\mathrm{int}}
&= \frac{\sum (E+\lambda_{\mathrm{int}}A)e}
{\sum(E+\lambda_{\mathrm{int}}A)},\\
\mathcal{L}_{\mathrm{video}}
&= \alpha\mathcal{L}_{\mathrm{env}}^{\mathrm{int}}
+(1-\alpha)\mathcal{L}_{\mathrm{robot}},
\end{aligned}
\label{eq:interaction_video_loss}
\end{equation}
where $A$ is a Gaussian-weighted prior restricted to the environment ($0\leq A\leq E$), and $\mathcal{L}_{\mathrm{robot}}$ supervises instrument regions. We set $\alpha=0.75$ when both regions are valid and use uniform supervision when geometry is unavailable. This objective emphasizes potential tissue responses without explicit contact annotations.

The overall objective is $\mathcal{L}=\mathcal{L}_{\mathrm{act}}+\mathcal{L}_{\mathrm{video}}+\mathcal{L}_{\mathrm{state}}+\mathcal{L}_{\mathrm{aux}}$, where $\mathcal{L}_{\mathrm{act}}$ denotes action flow matching. For $\mathcal{L}_{\mathrm{state}}$, selected interaction states are decoded into RGB frames through a learnable readout and frozen VAE decoder. Masked Charbonnier losses supervise RGB reconstruction and temporal-difference matching outside instrument regions, with a temporal weight of 0.5. Finally, $\mathcal{L}_{\mathrm{aux}}$ includes feature-transport regularization, state--video consistency, and forward-kinematics supervision.

\begin{table*}[t]
\centering
\caption{Video-generation results on dVRK-Chole5.2k under three action-conditioning settings: no future-action conditioning, model-predicted actions ($^{*}$), and ground-truth actions ($^{\ddagger}$). Backbone sizes denote the nominal parameter counts of the pretrained video backbones in billions. Bold and underlined values indicate the best and second-best results within each setting, respectively.}
\label{tab:main_results}
\resizebox{\textwidth}{!}{
\begin{tabular}{l|c|ccccccc}
\toprule
Methods & Backbone (B) & SSIM$\uparrow$ & PSNR (dB)$\uparrow$ & LPIPS$\downarrow$ & FVD$\downarrow$ & FID$\downarrow$ & Flow EPE$\downarrow$ & Flow MR$\rightarrow 1$ \\
\midrule

\rowcolor{purple!18}
\multicolumn{9}{c}{\textit{Action-Free Video Generation}} \\
\midrule

CogVideoX1.5~\cite{yang2025cogvideox} & 5 & 0.665 & 15.482 & 0.477 & 345.436 & 42.072 & 1.309 & 0.649 \\

HunyuanVideo-1.5~\cite{wu2025hunyuanvideo} & 8.3 & \underline{0.732} & 17.462 & 0.348 & \textbf{104.288} & \underline{24.091} & \underline{0.785} & \underline{0.694} \\

Cosmos-H-Predict~\cite{he2025cosmos} & 2 & 0.713 & \underline{17.492} & \underline{0.339} & 147.821 & 30.445 & 0.892 & 0.682 \\

Wan2.2~\cite{wan2025wan} & 5 & \textbf{0.737} & \textbf{18.510} & \textbf{0.297} & \underline{109.299} & \textbf{22.314} & \textbf{0.736} & \textbf{0.857} \\

\midrule
\rowcolor{green!15}
\multicolumn{9}{c}{\textit{Predicted-Action Video Generation}} \\
\midrule

DreamZero$^{*}$~\cite{ye2026world} & 5 & 0.615 & 17.711 & 0.408 & 667.806 & 81.548 & 0.963 & \underline{0.792} \\

OpenWAM$^{*}$~\cite{wang2026openwam} & 5 & 0.704 & 18.819 & 0.320 & 457.854 & 43.443 & 0.892 & 0.740 \\

X-WAM$^{*}$~\cite{guo2026unified} & 5 & \underline{0.752} & \underline{20.394} & \underline{0.232} & \underline{182.263} & \underline{24.738} & \underline{0.816} & 0.703 \\

KineState-SWAM$^{*}$ (Ours) & 5 & \textbf{0.773} & \textbf{21.686} & \textbf{0.216} & \textbf{79.874} & \textbf{14.369} & \textbf{0.744} & \textbf{0.820} \\

\midrule
\rowcolor{orange!20}
\multicolumn{9}{c}{\textit{Ground-Truth-Action Video Generation}} \\
\midrule

GigaWorld-1 Pro$^{\ddagger}$~\cite{team2026gigaworld} & 5 & 0.515 & 16.094 & 0.532 & 566.044 & 97.071 & 0.873 & \underline{0.888} \\

OpenWAM$^{\ddagger}$~\cite{wang2026openwam} & 5 & 0.721 & 18.941 & 0.301 & 336.505 & 37.821 & 0.823 & 0.783 \\

Cosmos-H-Dreams$^{\ddagger}$~\cite{tejero2026nvidia} & 2 & \underline{0.793} & 19.781 & 0.346 & 182.822 & 57.759 & 0.683 & 0.355 \\

DWM$^{\ddagger}$~\cite{kim2026dexterous} & 5 & 0.726 & \underline{20.118} & 0.255 & 235.605 & 45.642 & 0.794 & 0.795 \\

Kinema4D$^{\ddagger}$~\cite{xu2026kinema4d} & 5 & 0.748 & 19.162 & \underline{0.196} & \underline{155.791} & \underline{18.424} & \underline{0.676} & 0.861 \\

KineState-SWAM$^{\ddagger}$ (Ours) & 5 & \textbf{0.819} & \textbf{23.732} & \textbf{0.179} & \textbf{63.550} & \textbf{12.587} & \textbf{0.607} & \textbf{0.916} \\

\bottomrule
\end{tabular}
}
\vskip -0.1in
\end{table*}

\begin{table}[t]
\centering
\caption{Action prediction on dVRK-Chole5.2k. Joint nMAE measures normalized joint-space error, while tip ADE, tip FDE, and rotation error are evaluated through forward kinematics.}
\label{tab:action_results}

\resizebox{\columnwidth}{!}{
\begin{tabular}{l|cccc}
\toprule
Methods
& Joint nMAE$\downarrow$
& Tip ADE (mm)$\downarrow$
& Tip FDE (mm)$\downarrow$
& Rot. ($^\circ$)$\downarrow$ \\
\midrule

DreamZero
& 0.771 & 28.561 & 29.861 & 24.076 \\

OpenWAM
& 0.583 & 24.828 & 28.022 & 24.187 \\

X-WAM
& \underline{0.308}
& \underline{8.490}
& \underline{12.071}
& \underline{12.522} \\

\midrule

KineState-SWAM (Ours)
& \textbf{0.201}
& \textbf{5.416}
& \textbf{7.267}
& \textbf{7.765} \\

\bottomrule
\end{tabular}
}
 \vskip -0.1in
\end{table}

\begin{figure*}[t]
    \centering
    \includegraphics[width=\textwidth]{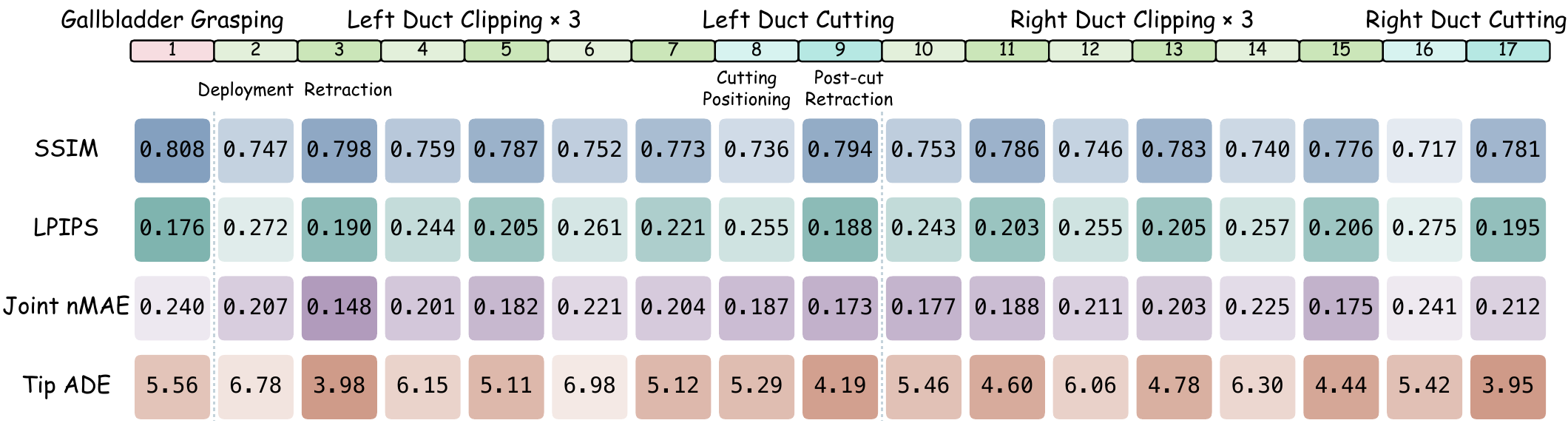}
   \caption{
    Phase-wise evaluation of KineState-SWAM across the 17 cholecystectomy subtasks in dVRK-Chole5.2k. The upper timeline indicates the procedural sequence. All evaluation metrics are computed from the model-predicted joint trajectories and their corresponding generated videos, without ground-truth action conditioning.
}
    \label{fig:phase_results}
         \vskip -0.1in
\end{figure*}

\section{Experimental Setup}
\label{sec:exp_setup}

\noindent\textbf{Dataset.} We construct dVRK-Chole5.2k from the ImitateCholec dataset~\cite{hansen2026imitatecholec} through quality screening and clip selection, including 5,255 videos at 30 FPS across 17 clipping-and-cutting subtasks. The training set contains 4,405 videos (583,040 frames), while the test set comprises 850 videos (110,040 frames) from independent surgical recordings. All methods are evaluated on the same test set.

\noindent\textbf{Baselines and metrics.} We compare KineState-SWAM against three categories of methods: (1) action-free video generation, including CogVideoX~\cite{yang2025cogvideox}, HunyuanVideo~\cite{wu2025hunyuanvideo}, Cosmos-H-Predict~\cite{he2025cosmos}, and Wan2.2~\cite{wan2025wan}; (2) predicted-action generation, including OpenWAM~\cite{wang2026openwam}, DreamZero~\cite{ye2026world}, and X-WAM~\cite{guo2026unified}; and (3) ground-truth-action-conditioned generation, including OpenWAM~\cite{wang2026openwam}, Cosmos-H-Dreams~\cite{tejero2026nvidia}, DWM~\cite{kim2026dexterous}, Kinema4D~\cite{xu2026kinema4d}, and GigaWorld-1~\cite{team2026gigaworld}. We evaluate video fidelity using SSIM, PSNR, LPIPS, FVD, and FID, and motion consistency using optical-flow endpoint error (EPE) and flow magnitude ratio (MR). Action prediction is assessed using normalized joint mean absolute error (nMAE), end-effector average displacement error (ADE), final displacement error (FDE), and rotation error (Rot. ($^\circ$)).

\begin{figure*}[t]
    \centering
    \includegraphics[width=\textwidth]{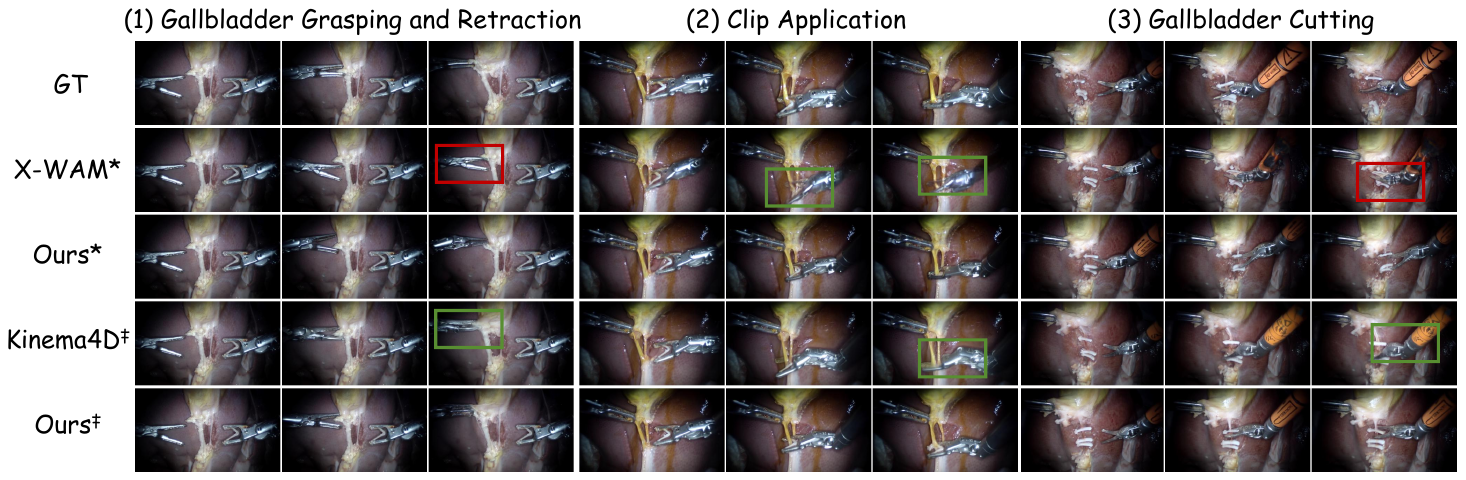}
   \caption{
   Qualitative comparison on dVRK-Chole5.2k for grasping and retraction, clip application, and cutting. Frames follow temporal order within each example, with ground truth in the top row. $^{*}$ and $^{\ddagger}$ denote predicted-action and ground-truth-action settings, respectively.
  Colored boxes highlight discrepancies in instrument configuration and local scene structure.
}
    \label{fig:qualitative_results}
         \vskip -0.1in
\end{figure*}

\noindent\textbf{Implementation details.}
We use Wan2.2-TI2V-5B as the video backbone, freeze its pretrained weights and VAE, and optimize the proposed modules and rank-32 LoRA adapters initialized from a surgically adapted checkpoint. All methods are LoRA-fine-tuned for approximately 6,000 steps on the same training set. We use AdamW with a global batch size of 8 and learning rates of $10^{-4}$ and $10^{-5}$ for newly introduced modules and LoRA, respectively. Training is conducted on 4 NVIDIA H100 80GB GPUs. During inference, the Action Policy predicts 16-frame trajectories, with interaction states updated every four frames. We generate 48 future frames per window at $832\times480$, using 10 action and 50 video sampling steps. Robot URDFs and camera calibration parameters are provided by the dataset.

\begin{figure}[t]
    \centering
    \includegraphics[width=\columnwidth]{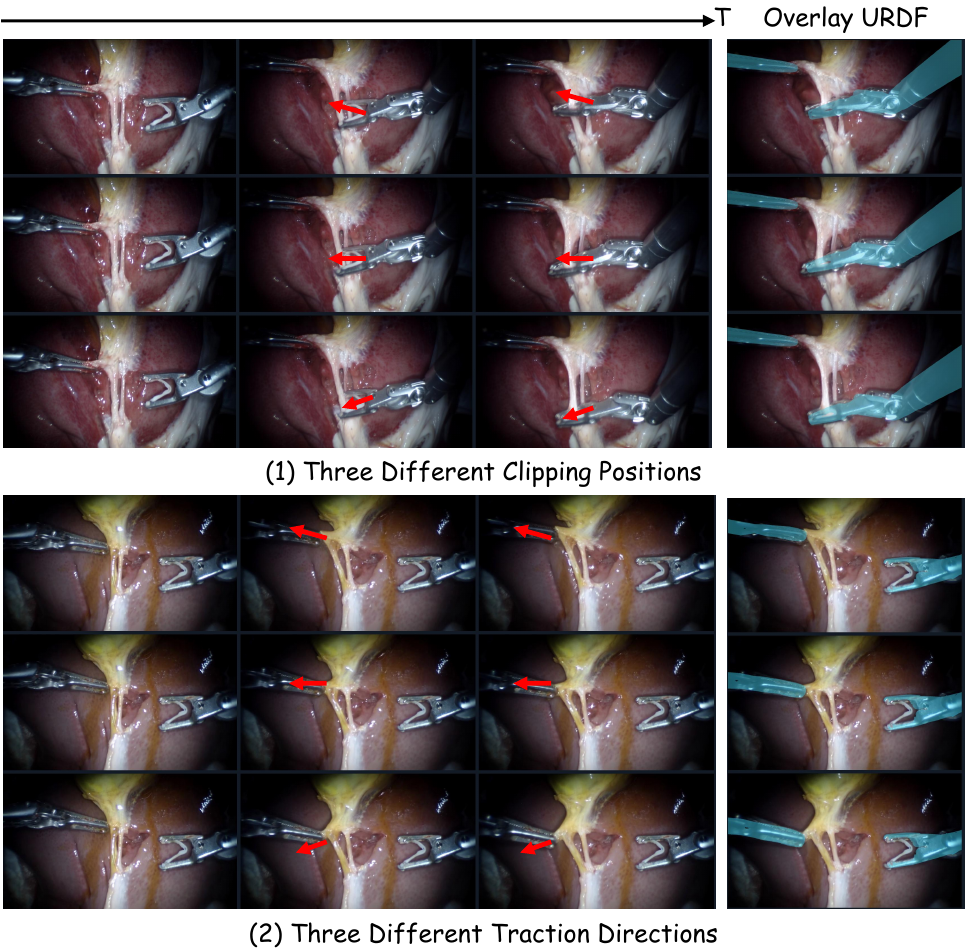}
    \caption{
    Action-conditioned video generation for three clipping positions and three traction directions.  Red arrows indicate the varied instrument motions, and the rightmost column overlays the corresponding URDF geometry for visual comparison.
    }
    \label{fig:action_control}
    \vskip -0.2in
\end{figure}

\section{Results and Analysis}

\begin{figure*}[t]
    \centering
    \includegraphics[width=\textwidth]{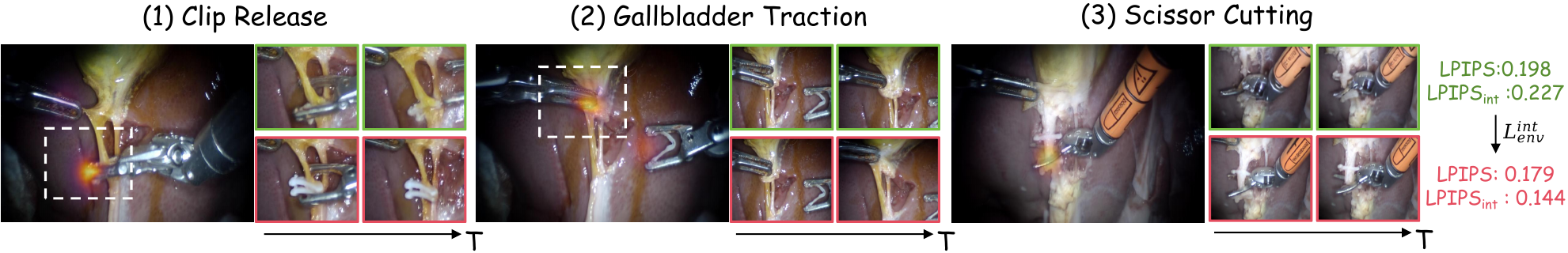}
     \caption{Ablation of interaction-aware supervision. Green (top) and red (bottom) crops show predictions without and with interaction-aware weighting, respectively. $\mathrm{LPIPS}_{\mathrm{int}}$ measures interaction-region perceptual error.}
    \label{fig:interaction_loss}
         \vskip -0.15in
\end{figure*}

\begin{figure}[t]
    \centering
    \includegraphics[width=\columnwidth]{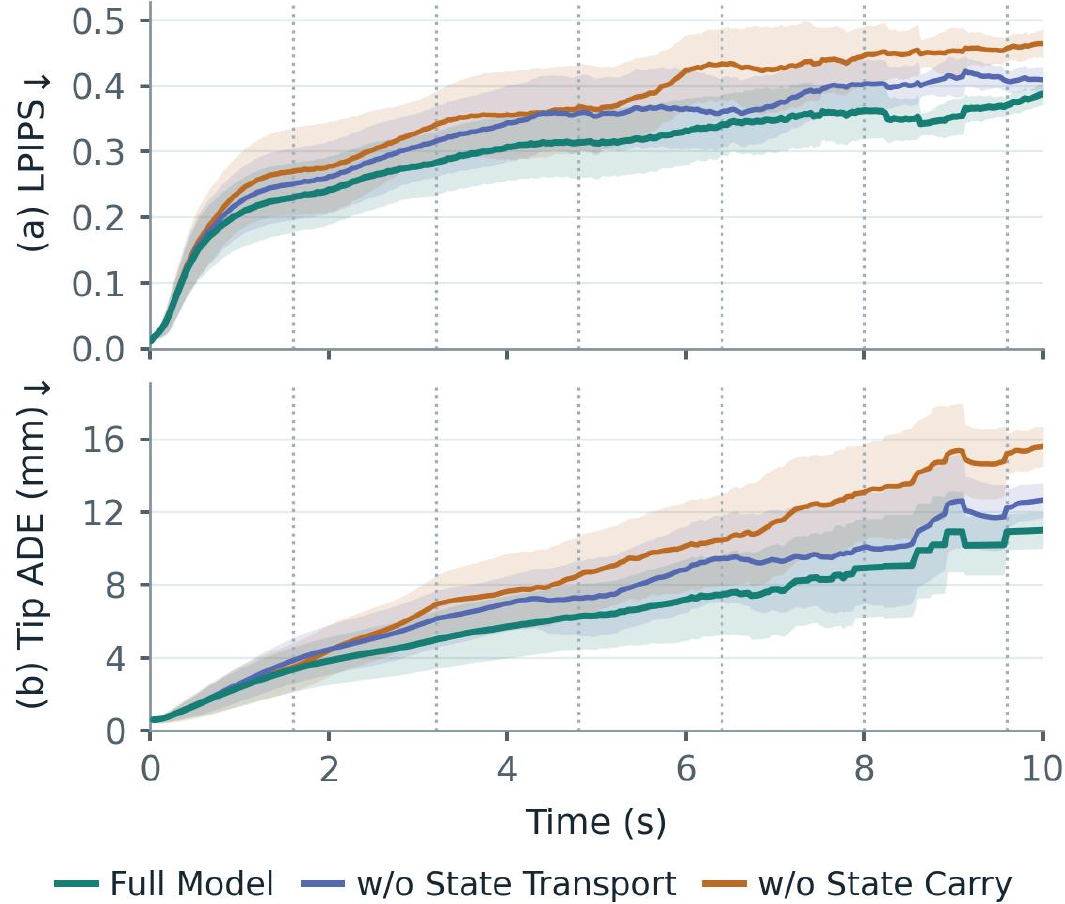}
    \caption{
    Prediction errors over time on the test set for the full model and variants without state transport or state carry. Curves and shaded regions indicate the mean and standard deviation across test sequences. Evaluation is performed on the full test set with variable-length sequences, averaging results at each prediction horizon over sequences with available ground-truth frames.
    }
    \label{fig:long_horizon}
    \vskip -0.2in
\end{figure}

\subsection{Quantitative Analysis}
\label{sec:quantitative_results}

\noindent\textbf{Video generation evaluation.} Table~\ref{tab:main_results} compares KineState-SWAM with video generation and world--action models under action-free, predicted-action, and ground-truth-action settings. With predicted actions, our model achieves the best results across all seven metrics within this setting. Compared with X-WAM, it improves SSIM from 0.752 to 0.773 and reduces FVD from 182.263 to 79.874 and FID from 24.738 to 14.369, corresponding to reductions of 56.2\% and 41.9\%, respectively. Flow EPE also decreases from 0.816 to 0.744, indicating improved motion agreement alongside higher visual fidelity. With GT actions, KineState-SWAM again leads the comparison, achieving 0.819 SSIM, 0.179 LPIPS, and 63.550 FVD. Relative to Kinema4D, it reduces FVD by 59.2\% and flow EPE by 10.2\%. These improvements with known future trajectories support the effectiveness of the proposed interaction-state and geometry conditioning for scene prediction.

\noindent\textbf{Action prediction evaluation.} As shown in Table~\ref{tab:action_results}, KineState-SWAM achieves the lowest error across all four action metrics. Compared with X-WAM, it reduces tip ADE and FDE by 36.2\% and 39.8\%, reaching 5.416\, mm and 7.267\, mm, respectively. Joint nMAE decreases from 0.308 to 0.201, while rotation error drops from 12.522$^\circ$ to 7.765$^\circ$. These consistent gains indicate improved trajectory accuracy in both joint configurations and end-effector poses.

\noindent\textbf{Subtask-wise evaluation.} Figure~\ref{fig:phase_results} jointly evaluates video fidelity and action accuracy across the 17 clipping-and-cutting subtasks. During left-duct clipping, retraction steps exhibit higher SSIM and lower LPIPS and tip ADE than the preceding clip-deployment steps. For example, between subtasks~2 and~3, LPIPS decreases from 0.272 to 0.190 and tip ADE from 6.78 to 3.98\,mm. The best-performing subtask also varies across metrics: gallbladder grasping achieves the highest SSIM (0.808), while subtasks~3 and~17 achieve the lowest joint nMAE (0.148) and tip ADE (3.95\,mm), respectively. These results highlight the varying challenges across surgical subtasks and the importance of jointly evaluating video fidelity and action accuracy.

\subsection{Qualitative Analysis}
\label{sec:qualitative_analysis}

\noindent\textbf{Comparison with baselines.}
Figure~\ref{fig:qualitative_results} compares generated sequences across grasping and retraction, clip application, and cutting. Under predicted actions, X-WAM exhibits inaccurate instrument configurations (red boxes) and texture distortions (green boxes). Even with ground-truth actions, Kinema4D shows localized appearance artifacts and less faithful tissue responses. In contrast, KineState-SWAM better preserves instrument textures and predicts interaction-induced tissue deformations more consistently with the reference sequences. These results highlight improved visual fidelity and more coherent instrument--tissue interactions under both action-conditioning settings.

\noindent\textbf{Action-conditioned generation.} Figure~\ref{fig:action_control} examines how generated scenes respond to instrument trajectories, including three clipping positions and three traction directions. The clipping sequences place the instrument at distinct target locations, while the traction sequences exhibit different pulling directions accompanied by changes in nearby tissue. The URDF overlays show alignment between the generated instruments and their prescribed geometric configurations. Together, these examples illustrate controllable instrument motion and action-dependent scene evolution, supporting the use of explicit kinematic guidance to connect robot trajectories with their visual consequences.

\begin{table}[t]
\centering
\caption{
Comparison of different control representations for surgical video generation using ground-truth actions.
}
\label{tab:control_representation}
\resizebox{\columnwidth}{!}{
\begin{tabular}{lccccc}
\toprule
\textbf{Control Representation}
& \textbf{SSIM} $\uparrow$
& \textbf{LPIPS} $\downarrow$
& \textbf{FVD} $\downarrow$
& \textbf{Flow EPE} $\downarrow$
& \textbf{Flow MR $\rightarrow 1$}  \\
\midrule
No Control (Wan2.2)
& 0.737 & 0.297 & 109.299 & 0.736 & 0.857 \\
Pose Control
& 0.752 & 0.278 & 105.642 & 0.721 & 0.872 \\
Mask Control
& 0.768 & 0.251 & 96.477 & 0.697 & 0.887 \\
URDF Render Control
& 0.790 & 0.215 & 81.509 & 0.651 & 0.902 \\
\midrule
\textbf{Ours}
& \textbf{0.819}
& \textbf{0.179}
& \textbf{63.550}
& \textbf{0.607}
& \textbf{0.916} \\
\bottomrule
\end{tabular}
}
 \vskip -0.1in
\end{table}

\begin{table}[t]
\centering
\caption{
Ablation of Causal Interaction Dynamics and the State--Geometry Adapter on dVRK-Chole5.2k. Variants remove or replace individual components of the full model.
}
\label{tab:component_ablation}
\resizebox{\columnwidth}{!}{
\begin{tabular}{lccccc}
\toprule
\textbf{Variant}
& \textbf{SSIM} $\uparrow$
& \textbf{LPIPS} $\downarrow$
& \textbf{FVD} $\downarrow$
& \textbf{Flow EPE} $\downarrow$
& \textbf{Flow MR} $\rightarrow 1$  \\
\midrule
\multicolumn{6}{c}{
\textcolor{purple!75!black}{
\textbf{Causal Interaction Dynamics}
}} \\
\midrule
w/o Explicit Surface Velocity
& 0.794 & 0.205 & 83.477 & 0.694 & 0.881 \\
w/o Geometric Bias
& 0.811 & 0.186 & 68.731 & 0.629 & 0.901 \\
w/o State Transport
& 0.801 & 0.197 & 78.908 & 0.682 & 0.892 \\
\midrule
\multicolumn{6}{c}{
\textcolor{orange!85!black}{
\textbf{State--Geometry Adapter}
}} \\
\midrule
Additive State Conditioning
& 0.809 & 0.191 & 76.541 & 0.667 & 0.874 \\
w/o Geometry Residual
& 0.791 & 0.209 & 85.115 & 0.693 & 0.899 \\
Global Geometry Residual ($P_R=\mathbf{1}$)
& 0.807 & 0.191 & 72.722 & 0.642 & 0.903 \\
\midrule
\textbf{Ours (Full)}
& \textbf{0.819}
& \textbf{0.179}
& \textbf{63.550}
& \textbf{0.607}
& \textbf{0.916} \\
\bottomrule
\end{tabular}
}
 \vskip -0.1in
\end{table}

\subsection{Ablation Studies}
\label{sec:ablations}

\noindent\textbf{Control representations.} Table~\ref{tab:control_representation} compares video generation with different control representations and uses ground-truth actions. URDF render control outperforms pose and mask control across all reported metrics, highlighting the value of spatially aligned instrument geometry. The full model further improves SSIM from 0.790 to 0.819 and LPIPS from 0.215 to 0.179, while reducing FVD from 81.509 to 63.550, a 22.0\% reduction. These results show that the complete interaction-state and conditioning design provides additional gains beyond URDF rendering alone.

\noindent\textbf{Causal Interaction Dynamics.} In Table~\ref{tab:component_ablation}, removing explicit surface velocity causes the largest degradation among the dynamics variants, increasing FVD from 63.550 to 83.477 and flow EPE from 0.607 to 0.694. Removing state transport raises these metrics to 78.908 and 0.682, respectively, while removing geometric bias also worsens all reported metrics. These results support combining explicit kinematic motion, geometry-guided attention, and motion-aligned feature transport for interaction-state prediction.

\noindent\textbf{State--Geometry Adapter.} In Table~\ref{tab:component_ablation}, replacing state modulation with additive conditioning increases FVD to 76.541 and moves flow MR from 0.916 to 0.874. Removing the geometry residual causes the largest FVD increase among the adapter variants, while applying it globally ($P_R=1$) increases FVD to 72.722 and flow EPE to 0.642. These results support combining spatial state modulation with localized geometric residuals for video generation.

\noindent\textbf{Long-horizon recurrent prediction.} 
Figure~\ref{fig:long_horizon} visualizes video and action prediction errors over time on the test set. The full model maintains lower LPIPS and tip ADE at longer prediction horizons than variants without state transport or state carry. These results support preserving and updating interaction history to improve subsequent action prediction and video generation.

\noindent\textbf{Interaction-aware supervision.} Figure~\ref{fig:interaction_loss} examines interaction-aware weighting in $\mathcal{L}_{\mathrm{env}}^{\mathrm{int}}$. To evaluate tissue responses independently of instrument appearance, $\mathrm{LPIPS}_{\mathrm{int}}$ is computed over interaction regions with URDF-derived instrument masks excluded. This weighting reduces LPIPS from 0.198 to 0.179 and $\mathrm{LPIPS}_{\mathrm{int}}$ from 0.227 to 0.144. These results demonstrate improved perceptual fidelity in tissue interaction regions while preserving video quality.

\FloatBarrier
\section{Conclusion}
\label{sec:conclusion}

In this work, we presented KineState-SWAM, a surgical world--action model that couples action prediction and video generation through a persistent, kinematics-driven interaction state. URDF-derived surface motion drives recurrent state updates, while the State--Geometry Adapter combines evolving interaction context with spatially aligned instrument geometry to guide video synthesis. Experiments demonstrate improved action accuracy and video quality compared with the evaluated baselines, with long-horizon ablations supporting the benefits of state retention and spatial transport during extended rollouts.
{
    \small
    \interlinepenalty=10000
    \bibliographystyle{ieeenat_fullname}
    \bibliography{main}
}
\end{document}